%% file: root.tex
\documentclass[letterpaper, 10 pt, conference]{ieeeconf}  
\IEEEoverridecommandlockouts                              %
\usepackage{caption}
\usepackage{cite}
\usepackage{amsmath,amssymb,amsfonts}
\usepackage{algorithmic}
\usepackage{graphicx}
\usepackage{textcomp}
\usepackage{xcolor}
\usepackage{subfigure}
\usepackage{multirow}
\usepackage{multicol}
\usepackage{diagbox}
\usepackage{float}
\usepackage{booktabs}
\usepackage{makecell}
\usepackage[colorlinks,linkcolor=blue]{hyperref}
\def\BibTeX{{\rm B\kern-.05em{\sc i\kern-.025em b}\kern-.08em
    T\kern-.1667em\lower.7ex\hbox{E}\kern-.125emX}}

\title{\LARGE \bf WHALES: A Multi-Agent Scheduling Dataset for \\Enhanced Cooperation in Autonomous Driving}

\author
 {Yinsong (Richard) Wang, Siwei Chen, Ziyi Song, Sheng Zhou*\thanks{* Corresponding Author: \texttt{sheng.zhou@tsinghua.edu.cn}}
 \thanks{
    The authors are with the Department of Electronic Engineering, Tsinghua University. Sheng Zhou is also with the State Key Laboratory of Intelligent Green Vehicle and Mobility, Tsinghua University. This work is sponsored in part by the project of Tsinghua University-Toyota Joint Research Center for AI Technology of Automated Vehicle.
 }
}

\begin{document}
\maketitle

\input{sections/00_abstract.tex}

\input{sections/01_intro}

\input{sections/02_related_work}

\input{sections/03_whales}

\input{sections/04_experiments}

\input{sections/05_conclusion}

\bibliography{references}
\bibliographystyle{ieeetr}

\end{document}

%% file: sections/00_abstract.tex
\begin{abstract}
Cooperative perception research is hindered by the limited availability of datasets that capture the complexity of real-world Vehicle-to-Everything (V2X) interactions, particularly under dynamic communication constraints. To address this gap, we introduce WHALES (\textbf{W}ireless en\textbf{H}anced \textbf{A}utonomous vehicles with \textbf{L}arge number of \textbf{E}ngaged agent\textbf{S}), the first large-scale V2X dataset explicitly designed to benchmark communication-aware agent scheduling and scalable cooperative perception. WHALES introduces a new benchmark that enables state-of-the-art (SOTA) research in communication-aware cooperative perception, featuring an average of 8.4 cooperative agents per scene and 2.01 million annotated 3D objects across diverse traffic scenarios. It incorporates detailed communication metadata to emulate real-world communication bottlenecks, enabling rigorous evaluation of scheduling strategies. To further advance the field, we propose the Coverage-Aware Historical Scheduler (CAHS), a novel scheduling baseline that selects agents based on historical viewpoint coverage, improving perception performance over existing SOTA methods. WHALES bridges the gap between simulated and real-world V2X challenges, providing a robust framework for exploring perception-scheduling co-design, cross-data generalization, and scalability limits. The WHALES dataset and code are available at \href{github}{https://github.com/chensiweiTHU/WHALES}. 
\end{abstract}

%% file: sections/01_intro.tex
\section{Introduction}

Modern autonomous vehicles rely on multi-modal sensor arrays to perceive complex environments and inform real-time decisions \cite{yurtsever2020survey}. Despite significant advances in perception tasks such as 3D object detection\cite{liu2023bevfusion}, 4D occupancy prediction\cite{tian2024occ3d}, multi-object tracking\cite{meinhardt2022trackformer}, and end-to-end planning\cite{planning}, isolated perception systems remain fundamentally constrained by limited sensor range, occlusions, and latency. Cooperative perception addresses these challenges by enabling data sharing between agents via Vehicle-to-Everything (V2X) networks, aggregating multi-agent perspectives. However, the development of robust cooperative algorithms is hindered by the lack of datasets that capture the complexity of real-world V2X environments.

\begin{figure}[ht]
\centering
\subfigure[Intersection]{
\begin{minipage}{0.46\linewidth}
\centering
\includegraphics[width=\linewidth]{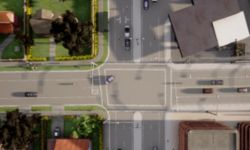}
\label{figure1:subfig:a}
\setlength{\abovecaptionskip}{-0.5cm}
\end{minipage}
}
\subfigure[T-junction]{
\begin{minipage}{0.46\linewidth}
\includegraphics[width=\linewidth]{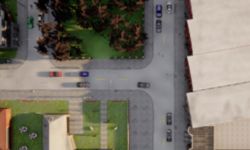}
\label{figure1:subfig:b}
\setlength{\abovecaptionskip}{-0.5cm}
\end{minipage}
}

\subfigure[Highway Ramp]{
\begin{minipage}{0.46\linewidth}
\includegraphics[width=\linewidth]{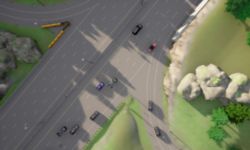}
\label{figure1:subfig:c}
\end{minipage}
}
\subfigure[5-way Intersection]{
\begin{minipage}{0.46\linewidth}
\includegraphics[width=\linewidth]{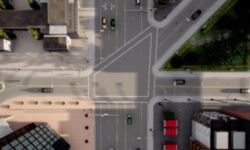}
\label{figure1:subfig:d}
\end{minipage}
}

\caption{{Representative road scenarios from the WHALES dataset, generated across multiple CARLA towns with varied agent and spawning strategies.}}
\label{figure1}
\end{figure}

\begin{figure*}[ht]
\centering
\subfigure[The BEV of a Single Frame]{\label{figure2:subfig:a}
\includegraphics[width=0.28\linewidth]{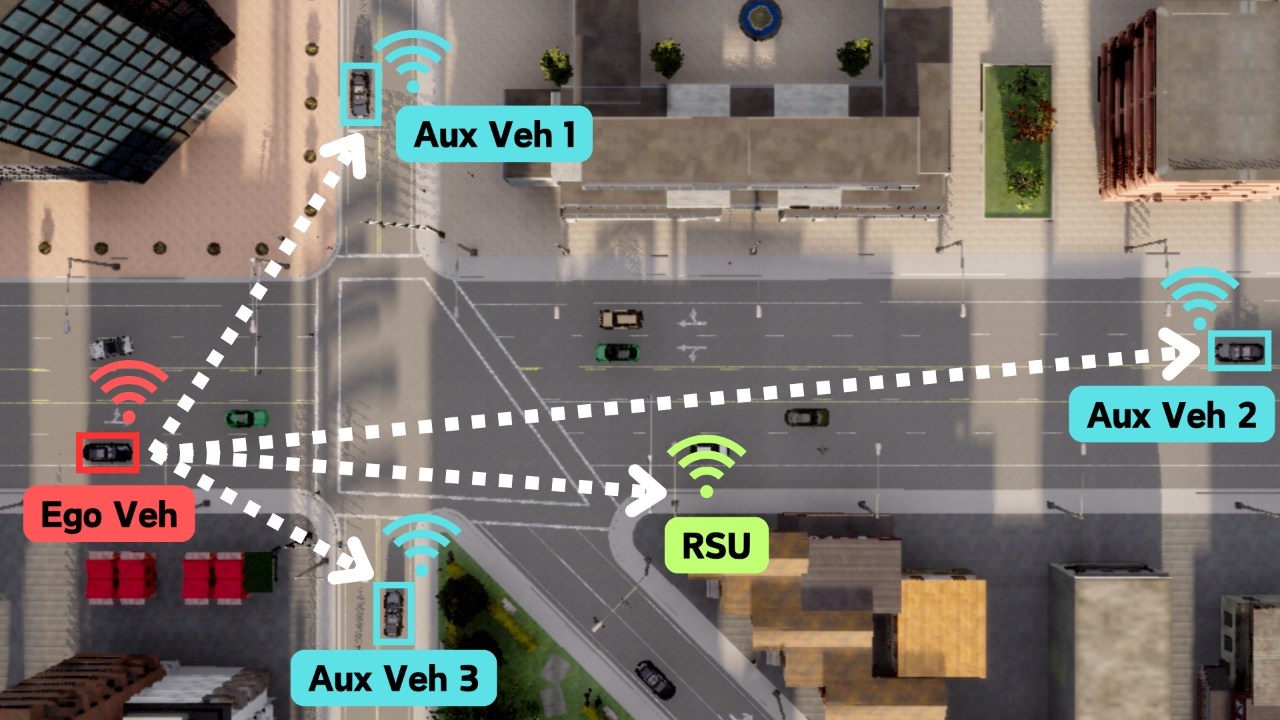}}
\subfigure[Aggregated Point Clouds in BEV]{\label{figure2:subfig:b}
\includegraphics[width=0.271\linewidth]{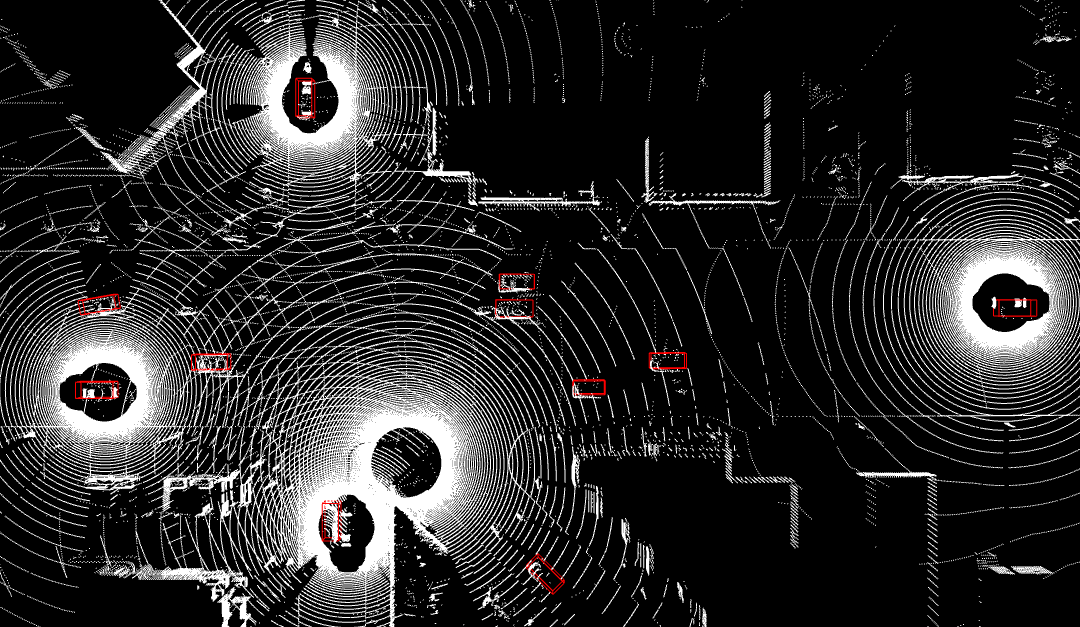}}
\subfigure[Agent's Front-camera Views of the Frame]{\label{figure2:subfig:c}
\includegraphics[width=0.335\linewidth]{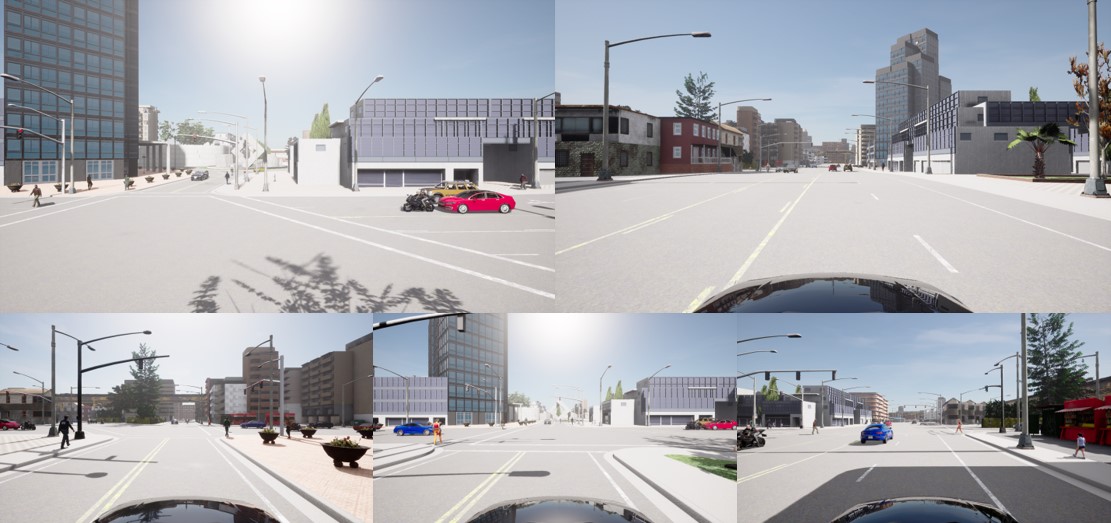}}
\caption{{Overview of the WHALES dataset. (a) Bird's-eye view (BEV) visualization of a representative frame, highlighting road topology and multi-agent interactions. (b) Aggregated LiDAR point clouds with 3D bounding box annotations (red) projected in BEV. (c) Front-camera views from multiple agents in the same frame.
}}
\label{figure2}
\end{figure*}

While several studies \cite{xu2022v2xvit,hu2022where2comm,lei2022syncnet,li2023lossy,hu2023coca} have focused on enhancing the robustness of cooperative perception (CP), they are evaluated on datasets that are often limited in scale. For instance, current benchmarks such as DAIR-V2X\cite{dairv2x} and DOLPHINS\cite{dolphins} are constrained to small-scale scenarios with only 2-3 agents per scene. This limitation stems from the significant computational cost of simulating dense multi-agent interactions. Although a few agents can provide broad sensor coverage, they are insufficient for resolving the complex, mutual occlusions common in dense urban traffic, a key challenge this work addresses. With a small number of collaborators, agent selection is minimal, and in contrast, with numerous potential partners and strict communication budgets, agent scheduling becomes indispensable.

Existing datasets do not adequately support research into this critical scheduling task, as they lack the necessary agent density. While simulated datasets like OPV2V\cite{opv2v} support Vehicle-to-Vehicle (V2V) cooperation, they omit infrastructure agents (V2I), lack adversarial conditions such as non-line-of-sight (NLOS) occlusions, and their static scenarios prevent research into adaptive scheduling. Furthermore, these datasets often lack the detailed communication metadata essential for evaluating robustness against the non-ideal factors explored in prior works \cite{lei2022syncnet}\cite{li2023lossy}\cite{lu2023robust}. These gaps limit progress toward communication-efficient, large-scale cooperative perception systems.

To overcome these barriers, we present \textbf{WHALES}, or \textbf{W}ireless en\textbf{H}anced \textbf{A}utonomous vehicles with a \textbf{L}arge number of \textbf{E}ngaged agent\textbf{S}, a comprehensive V2X dataset featuring 70K RGB images, 17K LiDAR frames and 2.01M 3D bounding box annotations. As shown in Fig. \ref{figure1}, WHALES models diverse road scenarios, including intersections, T-junctions, highway ramps, roundabouts, and 5-way intersections. Unlike prior autonomous driving datasets, WHALES integrates dynamic communication constraints, heterogeneous agents, including vehicles and infrastructure, and adversarial NLOS conditions. This enables detailed study of agent scheduling, model scalability, and robustness. Our benchmarks extend beyond conventional perception tasks by including communication-aware fusion and agent scheduling, promoting innovation in adaptive V2X systems.

Fig. \ref{figure2} provides an overview of a single frame from the WHALES dataset. In Fig. \ref{figure2:subfig:a}, the bird's-eye view (BEV) of the scenario is shown. Fig. \ref{figure2:subfig:b} visualizes the aggregated point cloud ranges from all agents with 3D bounding box annotations of objects, while the Fig. \ref{figure2:subfig:c} shows the front-camera views of these agents. The ego agent improves its detection of surrounding objects by scheduling cooperation with candidate agents.

Our contributions can be summarized as follows:
\begin{enumerate}
    \item We present the first simulated dataset enabling research on communication-constrained agent scheduling in V2X perception, featuring dynamic multi-agent scenarios, including adversarial NLOS conditions and occlusion patterns that necessitate robust cooperative reasoning.
    \item We propose the Coverage-Aware Historical Scheduler (CAHS) algorithm, a novel scheduling method that prioritizes agents based on their historical viewpoint coverage within the ego vehicle’s perception range. Through a handshake mechanism, CAHS dynamically selects candidates with maximal spatial relevance.
    \item We establish the benchmarks for 3D object detection and agent scheduling under communication constraints, and provide detailed analysis on one proposed and various existing baseline single-agent and multi-agent scheduling algorithms.
\end{enumerate}

%% file: sections/02_related_work.tex
\section{Related Work}

\subsection{Single-Agent Perception Datasets}
Large-scale datasets such as KITTI \cite{kitti}, nuScenes \cite{nuscenes}, and Waymo Open \cite{waymo} have significantly advanced research in 3D object detection and tracking by providing diverse, high-quality real-world sensor data. For motion forecasting and planning, datasets like Argoverse \cite{argoverse} and Argoverse2 \cite{argoverse2} offer detailed trajectory annotations and semantic maps, with nuPlan \cite{nuplan} introducing a closed-loop planning benchmark. However, these datasets are limited to single-agent settings, lacking the multi-agent interaction dynamics and collaborative data sharing essential for studying cooperative perception.

\subsection{Real-World V2X Perception Datasets}
To support collaborative perception, several Vehicle-to-Everything (V2X) datasets facilitate data exchange between vehicles and infrastructure. DAIR-V2X \cite{dairv2x} was the first large-scale real-world collaborative 3D detection dataset combining vehicle and roadside perspectives. V2V4Real \cite{v2v4real} improves multi-view perception through synchronized LiDAR and camera streams, while V2X-Seq \cite{v2xseq} provides sequential frames, traffic light states, and trajectory information for holistic scene understanding. TUMTraf‑V2X \cite{zimmer2024tumtrafv2x} focuses on real-world urban traffic by leveraging roadside cameras and LiDAR to capture long sequences for tracking and occlusion analysis across challenging scenarios such as intersections and roundabouts. V2X-Real \cite{xiang2024v2x} complements the ecosystem with densely labeled multi-agent LiDAR-camera data under both V2V and V2I configurations, offering synchronized streams across diverse urban environments. Nevertheless, these datasets are geographically limited and involve sparse agent interactions, restricting their utility for studying scalability and dense collaboration. 

\subsection{Simulated V2X Perception Datasets}

Simulation frameworks such as CARLA \cite{carla} enable scalable V2X dataset generation for cooperative tasks. OPV2V \cite{opv2v}, the first large-scale simulated dataset for Vehicle-to-Vehicle (V2V) perception, spans multiple towns and includes synchronized LiDAR-camera data. V2X-Sim \cite{v2xsim} adds Vehicle-to-Infrastructure (V2I) capabilities, and DOLPHINS \cite{dolphins} introduces dynamic weather conditions. However, these datasets average only 2-3 agents per scene, limiting the diversity of interaction topologies and hindering the evaluation of scalable scheduling policies. Moreover, their reliance on CARLA’s autopilot for agent control reduces behavioral diversity and constrains the simulation of complex multi-agent coordination scenarios.

\begin{table*}[t]
\setlength{\abovecaptionskip}{-0.5cm}
\caption{A Detailed Comparison Between WHALES and Existing Autonomous Driving Datasets}
\begin{center}
\begin{tabular}{lcccccccc} 
\\
\hline
\textbf{Dataset} & \textbf{Year} & \textbf{Source}  & \textbf{V2X} & \textbf{RGB images} & \textbf{LiDAR} & \textbf{3D boxes} & \textbf{Categories} & \textbf{Agent Number}\\

\hline
KITTI \cite{kitti} & 2012 & real & No & 15k & 15k & 200k & 8 & 1\\
nuScenes \cite{nuscenes} & 2019 & real & No & 1.4M & 400k & 1.4M & 23 & 1\\ 
DAIR-V2X \cite{dairv2x} & 2021 & real & V2V\&I & 39k & 39k & 464k & 10 & 2\\
V2X-Sim \cite{v2xsim} & 2021 & sim & V2V\&I & 0 & 10k & 26.6k & 1 & 2\\
OPV2V \cite{opv2v} & 2022 & sim & V2V & 44k & 11k & 230k & 1 & 2.9\\
DOLPHINS \cite{dolphins} & 2022 & sim & V2V\&I & 42k & 42k & 293k & 3 & 3\\
V2V4Real \cite{v2v4real} & 2023 & real & V2V & 40k & 20k & 240k & 5 & 2\\
TUMTraf‑V2X \cite{zimmer2024tumtrafv2x} & 2024 & real & V2V\&I & 5k & 2k & 30k & 8 & 3\\
V2X-Real \cite{xiang2024v2x} & 2024  & real & V2V\&I & 171k & 33k & 1.2M & 10 & 4 \\
\hline
\textbf{WHALES (Ours)} & 2025 & sim & V2V\&I & 70k & 17k & \textbf{2.01M} & 3 & \textbf{8.4}\\
\hline
\end{tabular}
\label{table1}
\end{center}
\end{table*}

\begin{table}[]
\centering
\small
\caption{Sensor specifications for CAVs and RSUs}
\vspace{-5pt}
\setlength{\tabcolsep}{4pt}
\begin{tabular}{l l}
\hline
\textbf{Sensors}    & \textbf{Details}                                \\ \hline
4x Camera  & RGB, Tesla: $1920 \times 1080$   \\ \hline
1x LiDAR & \begin{tabular}[c]{@{}l@{}}$64$~channels, $256k$ points per second, \\ $200m$ capturing range, $-40^\circ$ to $0^\circ$ \\ vertical FOV, 20Hz\end{tabular} \\ \hline
\end{tabular}
\vspace{-2mm}

\label{table:sensor-specs}
\vspace{-3mm}
\end{table}

\subsection{Scheduling Under Non-Ideal Conditions}
Recent works have explored model robustness under non-ideal V2X settings. The Mobility-Aware Sensor Scheduling (MASS) algorithm \cite{jia2023mass} leverages vehicle dynamics and Upper Confidence Bound (UCB) framework \cite{jia2022online} to enable decentralized scheduling and maximize perception gain. While MASS addresses latency and occlusion, it relies on fixed agent roles and does not scale beyond small agent clusters. Critically, no existing dataset or approach comprehensively supports the evaluation of scheduling under dense and dynamic agent interactions characteristic of real-world urban settings. Our work fills this critical gap by enabling scalable, communication-aware scheduling under realistic multi-agent conditions.

%% file: sections/03_whales.tex
\section{WHALES Dataset}
\begin{table*}[t]
\setlength{\abovecaptionskip}{-0.2cm}
\caption{Agent Categories and Configurations in the WHALES Dataset}
\begin{center}
\begin{tabular}{clcccc} 

\hline
\textbf{Agent Location} & \textbf{Agent Type} & \textbf{Sensor Suite} & \textbf{Control Strategy} & \textbf{Primary Role} & \textbf{Spawning Positions} \\
\hline

\multirow{2}{*}{On-Road} & Uncontrolled CAV & LiDAR $\times$ 1 + Camera $\times$ 4 & CARLA Autopilot & Perception & Random / Deterministic \\ 
\cline{2-6}
 & Controlled CAV & LiDAR $\times$ 1 + Camera $\times$ 4 & RL Policy & Perception \& Planning & Random / Deterministic \\
\hline
Roadside & Roadside Unit & LiDAR $\times$ 1 + Camera $\times$ 4 & RL Policy & Perception \& Planning & Static \\
\hline
Everywhere & Obstacle Agent & Not equipped & CARLA Autopilot & None & Random \\ 
\hline
\end{tabular}
\label{agent_config}
\end{center}
\end{table*}

\begin{figure*}[ht]
\centering
\subfigure[]{\label{dataset_distribution:subfig:a}
\includegraphics[width=0.23\linewidth]{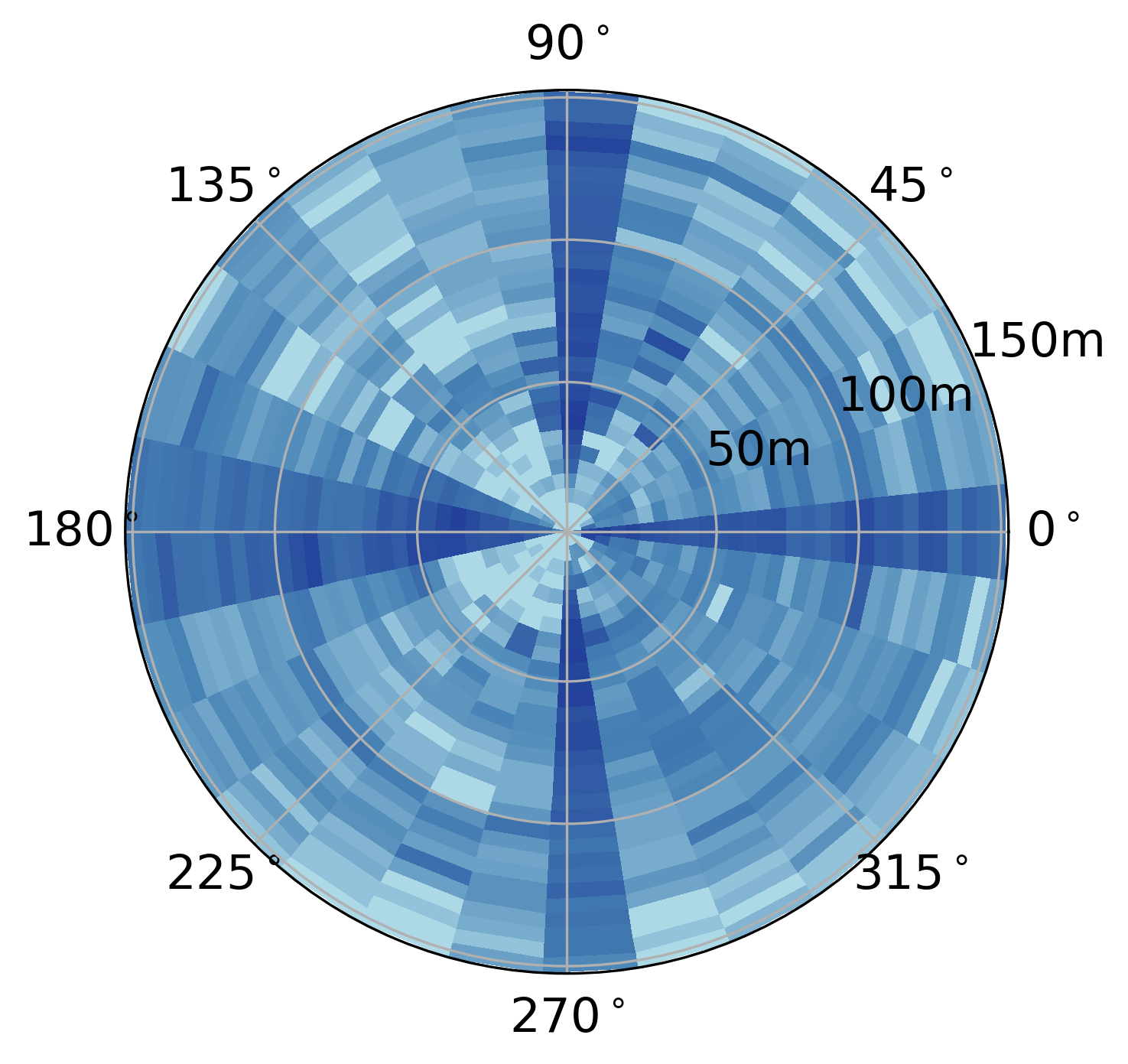}}
\hspace{1cm}
\subfigure[]{\label{dataset_distribution:subfig:b}
\includegraphics[width=0.23\linewidth]{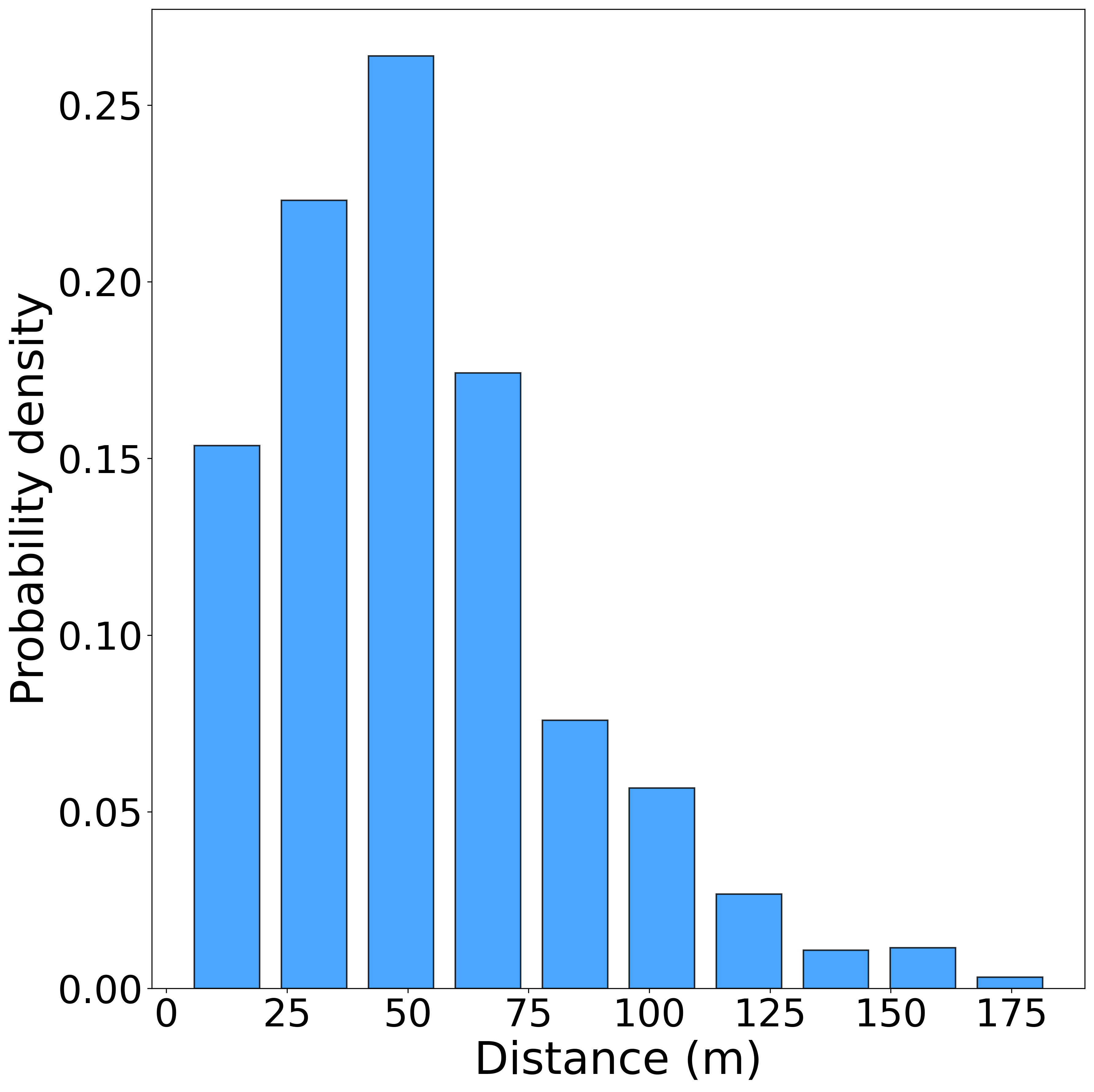}}
\hspace{1cm}
\subfigure[]{\label{dataset_distribution:subfig:c}
\includegraphics[width=0.23\linewidth]{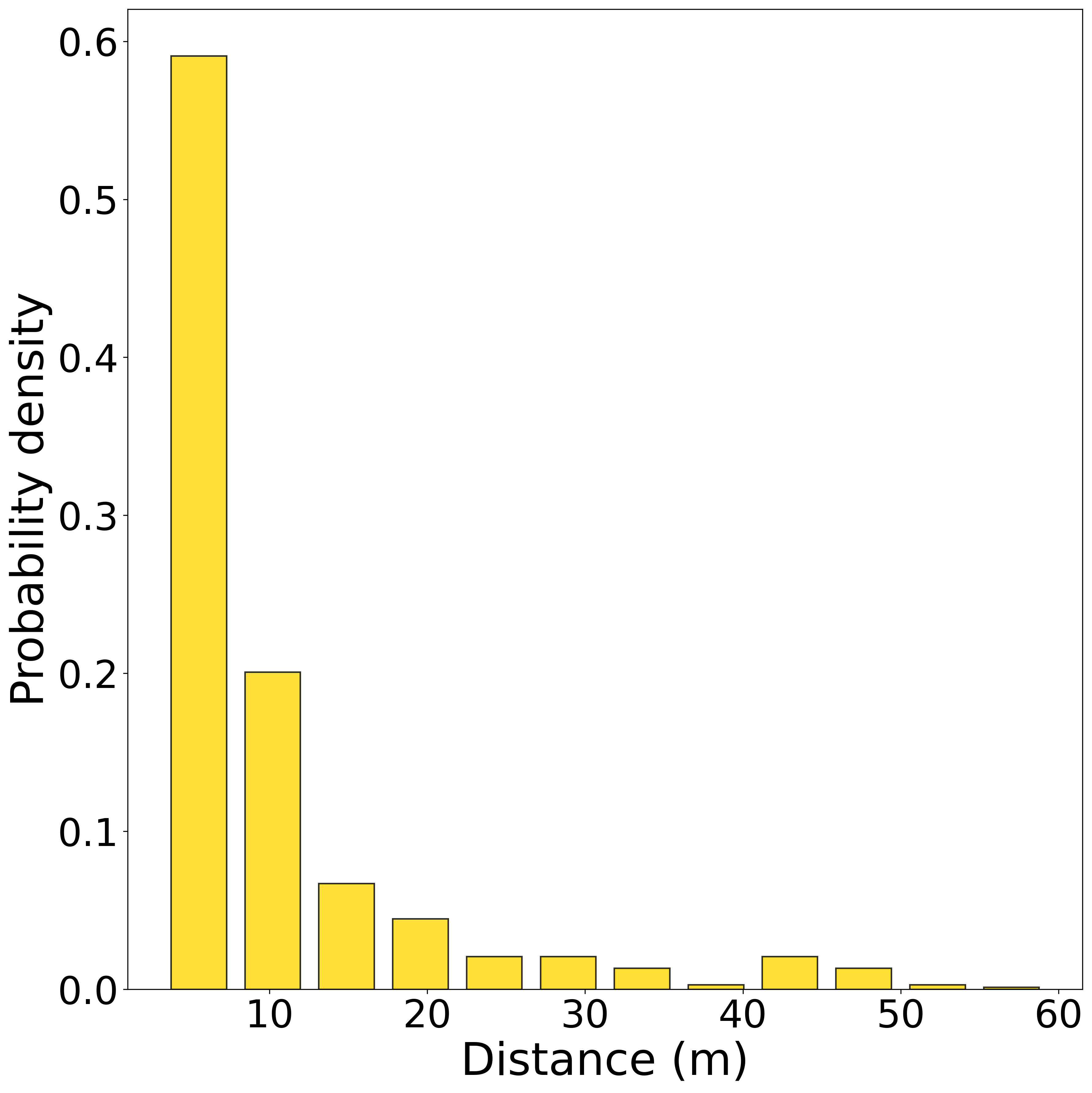}}
\caption{Spatial distribution characteristics in the WHALES dataset. (a) Orientation and position density of annotated 3D bounding boxes. (b) Distribution of distances between cooperative agents. (c) Distribution of minimum distances between agents and annotated objects.}
\label{dataset_distribution}
\end{figure*}

Table~\ref{table1} compares WHALES with representative cooperative perception benchmarks, highlighting key gaps in agent scalability and support for communication-aware scheduling that our dataset addresses. WHALES offers comprehensive support for both V2V and V2I interactions, along with detailed road context through densely populated multi-agent scenes. This benchmark facilitates rigorous evaluation of multi-agent scheduling algorithms under realistic and dynamic conditions. In our setup, a valid 3D bounding box is defined for any object within a 50-meter radius of an agent's viewpoint.

The computational cost of simulating dense multi-agent interactions increases non-linearly with agent density due to complex inter-agent dynamics. To address this, we implement a parallelized pipeline in which each active agent operates in a dedicated process, streaming raw sensor data to a synchronized queue. These processes are dynamically coordinated, and agent observations are aggregated into batches for efficient processing. A centralized expert model generates agent behaviors from the batched inputs. This architecture reduces per-agent simulation time to approximately 160 ms on an NVIDIA 4090 GPU, efficiently linearizing time complexity and enabling scalable dataset generation.

\subsection{Sensor and Agent Settings}
\label{AA}
The sensor and agent configurations in the WHALES dataset are detailed in Tables~\ref{table:sensor-specs} and~\ref{agent_config}. The dataset provides multi-modal sensory input to autonomous agents, balancing realism with computational efficiency. Four agent types are defined: (1) \textbf{Uncontrolled CAVs}, which follow predefined trajectories without external input; (2) \textbf{Controlled CAVs}, which are similarly equipped but controlled via user-defined planning algorithms; (3) \textbf{Roadside Units (RSUs)}, infrastructure agents outfitted with high-grade LiDARs and cameras; and (4) \textbf{Obstacle Agents}, which lack sensors and follow autonomous trajectories to represent ambient traffic participants. All CAVs and RSUs are equipped with a 64-beam LiDAR, four 1920 $\times$ 1080 RGB cameras, and V2X communication modules.

\subsection{Dataset Structure}
The dataset is hierarchically organized into scenes, frames, samples, and annotations. A scene represents a continuous simulation segment defined by specific environmental conditions and agent interactions. During simulation, sensor data is sampled at 0.1-second intervals, while full frames are recorded every 0.5 seconds to balance temporal resolution and storage efficiency. Each scene includes comprehensive simulation metadata and two synchronized video streams: one capturing the front-camera perspectives of all agents, and the other providing an encoded BEV map for holistic visualization, following the methodology of \cite{roach}. 

In every frame, the object class, 3D location, rotation, velocity, and unique track ID are recorded for all agents using a shared global coordinate system. To fully support cooperative perception tasks and ego-centric learning, WHALES cyclically assigns each agent the role of the ego vehicle within every frame. This design choice yields $N$ distinct samples per frame in scenarios with $N$ agents, capturing diverse ego-centric viewpoints. Annotations are generated by projecting global bounding boxes into each ego vehicle's local coordinate frame and filtering objects based on the perceptual range to better reflect real-world sensor limitations.

\subsection{Dataset Configuration System}
The dataset is generated using a modified version of the CARLA simulator\cite{carla}, integrated with a reinforcement learning (RL) framework from\cite{roach} and base environment configurations from\cite{opencda}. To simulate realistic cooperative scenarios, agents are dynamically spawned within rectangular regions centered on key traffic zones. This setup enables dense multi-agent interactions while preserving environmental plausibility. 

\subsection{Data Analysis}
Figure~\ref{dataset_distribution} illustrates the spatial properties of annotated bounding boxes in WHALES. As shown in Fig.~\ref{dataset_distribution:subfig:a}, object orientations are closely aligned with road topology, with most objects facing directions parallel or perpendicular to the ego vehicle’s heading. This results in prominent density peaks at 0°, 90°, 180°, and 270°.

Figure~\ref{dataset_distribution:subfig:b} depicts cooperative agent proximity, with most agents located within 50 meters of each other, allowing for complementary viewpoint coverage. A smaller subset is positioned beyond 100 meters to model long-range perception. Fig.~\ref{dataset_distribution:subfig:c} quantifies object proximity, showing that 88\% of objects fall within 20 meters of the nearest agent, highlighting the frequency of occlusions in dense urban environments.

Our annotation protocol follows the nuScenes benchmark \cite{nuscenes}, with all labels generated using ground-truth outputs from CARLA \cite{carla}. Annotations include all objects within a 50-meter valid detection range, including occluded instances, to reflect realistic perceptual challenges. In addition to standard 3D bounding boxes, we introduce cooperative-aware metadata such as inter-agent occlusion matrices and distance graphs to support the development of scheduling algorithms.

\begin{table*}[t]
\setlength{\abovecaptionskip}{-0.2cm}

\caption{Stand-alone 3D Object Detection Benchmark (50m/100m)}
\label{standalone_50/100m}
\begin{center}

\begin{tabular}{lcccc|ccccc}
\hline
\textbf{Method} & $\text{AP}_{Veh}\uparrow$ & $\text{AP}_{Ped}\uparrow$ & $\text{AP}_{Cyc}\uparrow$ & mAP$\uparrow$ & mATE$\downarrow$ & mASE$\downarrow$ & mAOE$\downarrow$ & mAVE$\downarrow$ & NDS$\uparrow$ \\
\hline

PointPillars \cite{lang2019pointpillars} & \textbf{67.1}/41.5 & 38.0/6.3 & \textbf{37.3}/11.6 & 47.5/19.8 & 0.117/0.247 & 0.876/0.880 & \textbf{1.069}/\textbf{1.126} & 1.260/1.625 & 33.8/18.6 \\
SECOND \cite{yan2018second} & 58.5/38.8 & 27.1/12.1 & 24.1/\textbf{12.9} & 36.6/21.2 & 0.106/0.156 & 0.875/0.878 & 1.748/1.729 & \textbf{1.005}/\textbf{1.256} & 28.5/20.3 \\
RegNet \cite{radosavovic2020designing} & 66.9/\textbf{42.3} & 38.7/8.4 & 32.9/11.7 & 46.2/20.8 & 0.119/0.240 & \textbf{0.874}/0.881 & 1.079/1.158 & 1.231/1.421 & 33.2/19.2 \\
VoxelNeXt \cite{chen2023voxenext} & 64.7/\textbf{42.3} & \textbf{52.2}/\textbf{27.4} & 35.9/9.0 & \textbf{50.9}/\textbf{26.2} & \textbf{0.075}/\textbf{0.142} & 0.877/\textbf{0.877} & 1.212/1.147 & 1.133/1.348 & \textbf{36.0}/\textbf{22.9} \\
\hline
\end{tabular}
\end{center}
\end{table*}

\begin{table*}[htbp]
\setlength{\abovecaptionskip}{-0.2cm}
\caption{Cooperative 3D Object Detection Benchmark (50m/100m)}
\label{cooperative_50/100m}
\begin{center}

\begin{tabular}{lcccc|ccccc}
\hline

\textbf{Method} & $\text{AP}_{Veh}\uparrow$ & $\text{AP}_{Ped}\uparrow$ & $\text{AP}_{Cyc}\uparrow$ & mAP$\uparrow$ & mATE$\downarrow$ & mASE$\downarrow$ & mAOE$\downarrow$ & mAVE$\downarrow$ & NDS$\uparrow$ \\

\hline

No Fusion & 67.1/41.5 & 38.0/6.3 & 37.3/11.6 & 47.5/19.8 & 0.117/0.247 & 0.876/0.880 & 1.069/\textbf{1.126} & \textbf{1.260}/1.625 & 33.8/18.6\\
\hline
F-Cooper & \textbf{75.4}/\textbf{52.8} & 50.1/9.1 & 44.7/20.4 & 56.8/27.4 &0.117/0.205&\textbf{0.874}/0.879&1.074/1.206&1.358/\textbf{1.449}& 38.5/22.9 \\

Raw-level Fusion & 71.3/48.9 & 38.1/8.5 & 40.7/16.3 & 50.0/24.6 & 0.135/0.242 & 0.875/0.882 & \textbf{1.062}/1.242 &1.308/1.469&34.9/21.1 \\

VoxelNeXt & 71.5/50.6 & \textbf{60.1}/\textbf{35.4} & \textbf{47.6}/\textbf{21.9} & \textbf{59.7}/\textbf{35.9} &\textbf{0.085}/\textbf{0.159}&0.877/\textbf{0.878}&1.070/1.204&1.262/1.463& \textbf{40.2}/\textbf{27.6} \\
\hline
\end{tabular}
\end{center}
\end{table*}

\subsection{Supported Tasks}

WHALES supports both stand-alone 3D object detection and cooperative perception tasks. A central challenge in cooperative driving is agent scheduling—selecting informative collaborators from a large pool of candidates. This problem is uniquely addressable at scale by WHALES, as existing datasets with fewer agents do not adequately support evaluation of scheduling strategies.

Ground-truth agent behaviors and trajectories, generated by RL-based expert policies \cite{roach}, enable imitation learning and trajectory-based policy evaluation. To ensure diversity and quality, generated scenes are filtered based on three criteria: (1) cumulative rewards achieved by the RL expert, (2) simulation duration, and (3) the distribution of agent categories. This curation supports future extensions to downstream tasks such as motion forecasting and multi-agent planning.

%% file: sections/04_experiments.tex
\section{Experiments}

\subsection{Experimental Details}
Experiments were conducted on 8 NVIDIA GeForce RTX 4090 GPUs. The dataset was split into training and testing sets with an 80/20 ratio for each task. Models were trained for 24 epochs with a base learning rate of 0.001 and evaluated at detection ranges of 50m and 100m. To simulate real-world communication constraints, data transmission during cooperative tasks was capped at 2MB per frame.

\subsection{Stand-alone 3D Object Detection}
We use the MMDetection3D framework\cite{MMDetection3D}, following experimental settings developed for the nuScenes dataset\cite{nuscenes}. Three representative architectures are evaluated: PointPillars\cite{lang2019pointpillars}, RegNet\cite{radosavovic2020designing} and SECOND\cite{yan2018second}, with a focus on detection performance at 50m and 100m ranges. Data augmentation strategies replicate those employed in the nuScenes\cite{nuscenes} training pipeline.

We adopt benchmark metrics consistent with nuScenes \cite{nuscenes}, reporting class-specific average precision (AP) for vehicles ($\text{AP}_{Veh}$), pedestrians ($\text{AP}_{Ped}$), and cyclists ($\text{AP}_{Cyc}$), as well as mean average precision (mAP) and the nuScenes Detection Score (NDS), which combines mAP with attributes like translation, scale, and orientation errors.

As shown in Table \ref{standalone_50/100m}, all models perform best on $\text{AP}_{Veh}$, due to the larger and more regular bounding box shapes of vehicles. Detection performance drops substantially at 100m, especially for pedestrians and cyclists, highlighting the limitations of stand-alone perception in long-range scenarios.

\subsection{Cooperative 3D Object Detection}
We benchmark cooperative perception using PointPillars\cite{lang2019pointpillars} as the baseline architecture, where agents share preprocessing layers, backbone networks, and detection heads. During training, each agent randomly selects one cooperative partner, while inference defaults to using the closest agent. This cooperative setup yields improved detection accuracy over stand-alone models. Furthermore, WHALES' larger pool of candidate agents enables further performance gains through agent scheduling.

Table~\ref{cooperative_50/100m} presents the results of cooperative 3D object detection experiments, with the No Fusion method, adapted from PointPillars\cite{lang2019pointpillars}, serving as the baseline. The cooperative variant of VoxelNeXt\cite{chen2023voxenext}, which employs sparse convolution for feature fusion, achieves state-of-the-art performance among single-level fusion methods. All cooperative approaches yield notable improvements over the baseline, underscoring the benefits of multi-agent collaboration.

Raw-level cooperative perception lags behind feature-level methods due to limited representational capacity, as it reuses the same backbone architecture as the No Fusion baseline. Compared to the baseline, F-Cooper\cite{chen2019fcooper} improves the mAP by 19.5\% and 38.4\% at 50m and 100m, respectively. VoxelNeXt\cite{chen2023voxenext} achieves even higher gains of 25.7\% and 81.3\%, highlighting the effectiveness of feature-level fusion in long-range detection.

\begin{table*}[t]
\setlength{\abovecaptionskip}{-0.2cm}

\caption{mAP Scores on 3D Object Detection using Different Scheduling Policies (50m/100m)}
\label{scheduling_50/100m}
\begin{center}
\begin{tabular}{l|ccccc}
\hline
\diagbox{\textbf{Inference}}{\textbf{Training}} & No Fusion & Closest Agent & Single Random & Multiple Random & Full Communication \\
\hline
\multicolumn{6}{c}{\textbf{Without Scheduling}} \\
\hline
No Fusion (Baseline) & 50.9/26.2 & 50.9/23.3 & 51.3/25.3 & 50.3/22.9 & 45.6/18.8 \\ 
\hline
\multicolumn{6}{c}{\textbf{Single-Agent Scheduling}} \\
\hline
Closest Agent & 39.9/20.3 & 58.4/30.2 & 58.3/32.6 & 57.3/30.5 & 55.4/10.8 \\
Single Random & 43.3/22.8 & 57.9/31.0 & 58.4/33.3 & 57.7/31.4 & 55.0/14.6 \\
MASS & 55.5/11.0 & 58.8/\textbf{33.7} & 58.9/34.0 & 57.3/32.3 & 54.1/27.4 \\
\textbf{CAHS (Proposed)} & \textbf{56.1}/\textbf{29.6} & \textbf{62.5}/31.7 & \textbf{62.7}/\textbf{35.9} & \textbf{58.3}/\textbf{32.6} & \textbf{59.9}/\textbf{31.0} \\
\hline
\multicolumn{6}{c}{\textbf{Multi-Agent Scheduling}} \\
\hline
Multiple Random & 34.5/\textbf{16.9} & 60.7/35.1 & 61.2/37.1 & 61.4/36.4 & 58.8/12.9 \\
Full Communication & 29.1/10.5 & 63.7/38.4 & 63.7/39.1 & \textbf{64.0}/\textbf{41.1} & 65.1/39.2 \\
MASS & \textbf{54.6}/13.4 & 64.9/39.7 & 65.0/40.5 & 63.7/40.4 & 63.5/36.4 \\
\textbf{CAHS (Proposed)} & 53.7/14.2 & \textbf{65.3}/\textbf{40.1} & \textbf{65.1}/\textbf{42.0} & 63.9/40.6 & \textbf{65.2}/\textbf{39.2} \\

\hline
\end{tabular}
\end{center}
\end{table*}
\subsection{Agent Scheduling}
\label{scheduling}

\begin{figure*}[ht]
\centering

\includegraphics[width=0.9\linewidth]{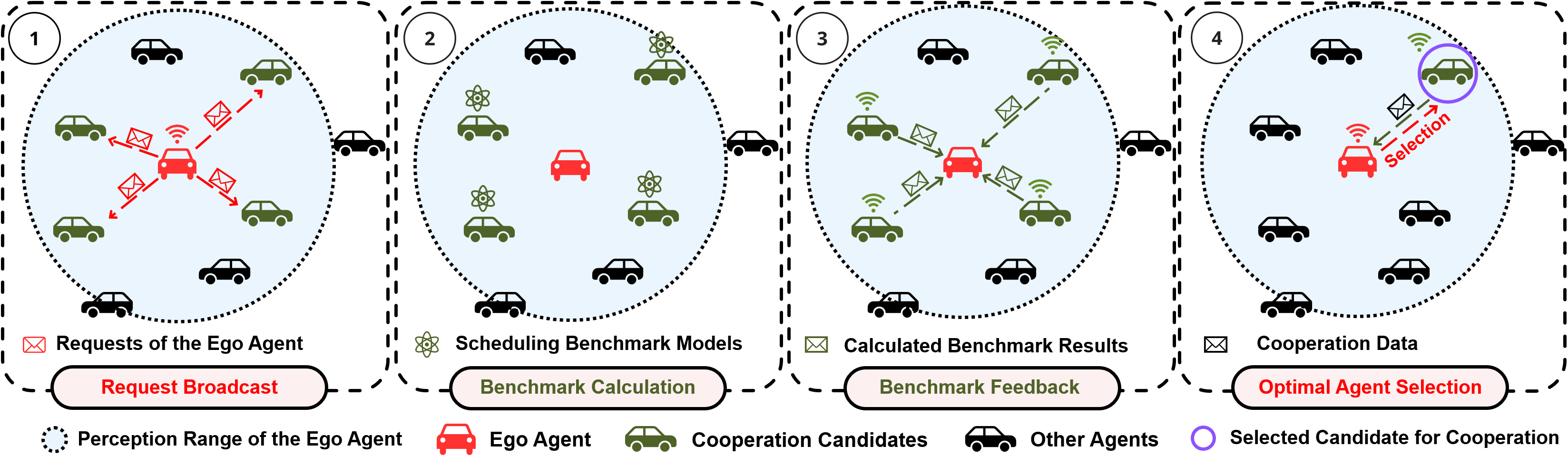}
\caption{{Visualization of the four stages of \textbf{agent scheduling}: \textbf{(1) Request Broadcast:} The ego agent broadcasts cooperation requests to all candidate. \textbf{(2) Benchmark Calculation:} Each candidate calculates scheduling benchmark relative to the ego's perception range. \textbf{(3) Benchmark Feedback:} Candidates send their benchmark scores back to the ego. \textbf{(4) Optimal Agent Selection:} The ego selects the optimal agent for collaboration based on the received benchmarks.}}
\label{scheduling_process}
\end{figure*}

As illustrated in Fig.~\ref{scheduling_process}, agent scheduling involves the ego agent selecting a candidate agent via unicast communication for raw-level fusion. In a scenario with \(N\) agents, there exist \(2^N\) possible combinations of sensory inputs. 
Exhaustively iterating over all configurations during training is computationally infeasible, making the design of efficient training and inference policies a key challenge in cooperative perception.
We evaluate a range of scheduling strategies using VoxelNeXt \cite{chen2023voxenext} as the model backbone. 

The \textbf{Full Communication} strategy allows the ego agent $E$ to aggregate bounding boxes from all $N$ candidate agents $\mathcal{C}=\{C_1,...,C_N\}$, representing an upper bound on perception performance:
\begin{equation}
    \mathcal{P}_{\text{full}}^t = \bigcup_{i=1}^N \mathcal{B}_i^t,
\end{equation}

where \( \mathcal{B}_i^t \) denotes the 3D bounding boxes detected by agent \( C_i \) at time \( t \). Conversely, the \textbf{No Fusion} method relies solely on ego agent's onboard sensors, defining a lower bound:
\begin{equation}
\mathcal{P}_{\text{ego}}^t = \mathcal{B}_E^t.
\end{equation}

A baseline method, \textbf{Closest Agent}, selects the nearest candidate \( C^*_{\text{closest}} \) based on the Euclidean distance \( d(\cdot) \):
\begin{equation}
C^*_{\text{closest}} = \underset{C_i \in \mathcal{C}}{\arg\min} \, d(\mathbf{x}_E^t, \mathbf{x}_i^t),
\end{equation}
where \( \mathbf{x}_E^t \) and \( \mathbf{x}_i^t \) are the coordinates of ego agent \( E \) and candidate \( C_i \) at time $t$.

We propose the \textbf{Coverage-Aware Historical Scheduler (CAHS)}, which prioritizes candidates based on historical utility. For candidate \( C_i \), its reward \( R_i^{t-1} \) is defined as the number of its previous detections intersecting with the ego agent’s past perception range \( \mathcal{R}_E^{t-1} \):
\begin{equation}
R_i^{t-1} = \sum_{b \in \mathcal{B}_i^{t-1}} \mathbb{I}\left(b \cap \mathcal{R}_E^{t-1} \neq \emptyset\right),
\end{equation}
where \( \mathbb{I}(\cdot) \) is the indicator function. The scheduling process includes:
\begin{enumerate}
    \item \( E \) broadcasts its location \( \mathbf{x}_E^t \) to all \( C_i \in \mathcal{C} \),
    \item Each \( C_i \) computes \( R_i^{t-1} \) and returns it to \( E \),
    \item \( E \) selects the optimal agent:
    \begin{equation}
    C^*_{\text{CAHS}} = \underset{C_i \in \mathcal{C}}{\arg\max} \, R_i^{t-1}.
    \end{equation}
\end{enumerate}

The results in Table~\ref{scheduling_50/100m} compare scheduling policies across two training and inference dimensions. We distinguish between single-agent scheduling where the ego agent selects one collaborator, and multi-agent scheduling, which allows fusion with multiple agents.

\textbf{Single-Agent Scheduling.} The proposed \textbf{CAHS} algorithm achieves the highest mAP across diverse training setups. When trained under the Single Random Agent policy, CAHS attains peak performance at both 50m and 100m detection ranges, surpassing the learning-based MASS algorithm~\cite{jia2023mass}, which suffers from memory constraints and limited context modeling.

\textbf{Multi-Agent Scheduling.} In the multi-agent setting, CAHS remains competitive and even outperforms the Full Communication and MASS~\cite{jia2023mass} baselines in several configurations. Notably, when trained with Full Communication, CAHS still achieves superior performance during inference, highlighting its robustness and communication efficiency.

\begin{table}[h]
\caption{mAP Scores on 3D Object Detection of CAHS with Different Number of Cooperative Agents (50m)}
    \centering
    \begin{tabular}{c c c c c c c c c}
        \toprule
        Agents & 1 & 2 &  3 & 4 & 5 & 6 & 7 & All\\
        \midrule
        mAP & 59.9 & 60.0 & 65.2 & \textbf{66.1} & 64.9 & 65.1 & 65.2 & 65.1\\
        \bottomrule
    \end{tabular}
    
    \label{map_cahs}
\end{table}

\textbf{Ablation on Scheduling.}
Our analysis reveals a nonlinear relationship between the number of cooperative agents and detection accuracy. As shown in Table~\ref{map_cahs}, when the CAHS algorithm is trained under the Full Communication policy, performance initially improves with additional agents, peaking at 66.1 mAP with four agents. However, beyond this point, accuracy plateaus and then declines, suggesting diminishing returns due to viewpoint redundancy and increased noise from conflicting observations. 

This trend is consistent with results in Table~\ref{scheduling_50/100m}, which demonstrates that full communication can degrade performance due to sensor overload and bandwidth constraints. These findings underscore the importance of dynamic agent scheduling to select a subset of spatially diverse and complementary agents, rather than aggregating data from all agents indiscriminately.

%% file: sections/05_conclusion.tex
\section{Conclusions}
We introduce \textbf{WHALES}, the first V2X dataset specifically designed to address scalability and communication-aware scheduling in cooperative perception. WHALES fills critical gaps in existing benchmarks by enabling scheduling evaluation under redundant agent coverage, realistic communication budgets, and scalable perception across densely populated scenes, capabilities that existing datasets do not jointly provide. To further advance agent scheduling, we propose the \textbf{Coverage-Aware Historical Scheduler (CAHS)}, a novel algorithm that balances viewpoint diversity and communication efficiency to improve perception performance.
